\documentclass[namedreferences,hyperref,optionalrh]{spr-sola}
\usepackage{graphicx}        
\usepackage{color}           

\newcommand{\aap}{{\it Astron. Astrophys.}}

\newcommand{\apj}{{\it Astrophys. J.}}
\newcommand{\apjl}{{\it Astrophys. J. Lett.}}
\newcommand{\apjs}{{\it Astrophys. J. Supp.}}

\newcommand{\grl}{{\it Geophys. Res. Lett.}}

\newcommand{\jgr}{{\it J. Geophys. Res.}}
\newcommand{\mnras}{{\it Mon. Not. R. Astron. Soc.}}

\newcommand{\solphys}{{\it Sol. Phys.}}

\chardef\us=`\_

\begin{document}

\begin{frontmatter}
\title{Prediction of Geoeffective CMEs Using SOHO Images and Deep Learning}
\author[addressref={aff1,aff2},corref,email={kalobaid@ksu.edu.sa}]
{\inits{K. A.}\fnm{Khalid A.}~\snm{Alobaid}} 
\author[addressref={aff1,aff3},corref,email={wangj@njit.edu}]
{\inits{J. T. L.}\fnm{Jason T. L.}~\snm{Wang}}
\author[addressref={aff1,aff4,aff5},corref,email={haimin.wang@njit.edu}]
{\inits{H.}\fnm{Haimin}~\snm{Wang}}
\author[addressref={aff1,aff4,aff5}]
{\inits{J.}\fnm{Ju}~\snm{Jing}}
\author[addressref={aff1,aff3}]
{\inits{Y.}\fnm{Yasser}~\snm{Abduallah}}
\author[addressref={aff1,aff3}]
{\inits{Z.}\fnm{Zhenduo}~\snm{Wang}}
\author[addressref=aff6]
{\inits{H.}\fnm{Hameedullah}~\snm{Farooki}}
\author[addressref=aff7]
{\inits{H.}\fnm{Huseyin}~\snm{Cavus}}
\author[addressref=aff5]
{\inits{V.}\fnm{Vasyl}~\snm{Yurchyshyn}}
\address[id=aff1]{Institute for Space Weather Sciences, New Jersey Institute of Technology, University Heights, Newark, NJ 07102, USA}
\address[id=aff2]{College of Applied Computer Sciences, King Saud University, Riyadh 11451, Saudi Arabia}
\address[id=aff3]{Department of Computer Science, New Jersey Institute of Technology, University Heights, Newark, NJ 07102, USA}
\address[id=aff4]{Center for Solar-Terrestrial Research, New Jersey Institute of Technology, University Heights, Newark, NJ 07102, USA}
\address[id=aff5]{Big Bear Solar Observatory, New Jersey Institute of Technology, 40386 North Shore Lane, Big Bear City, CA 92314, USA}
\address[id=aff6]{Department of Astrophysical Sciences,
Princeton University, Princeton, NJ 08544, USA}  
\address[id=aff7]{Department of Physics, Canakkale Onsekiz Mart University, 17110 Canakkale, Turkey}  

\runningauthor{Khalid A. Alobaid et al.}
\runningtitle{Prediction of Geoeffective CMEs}

\begin{abstract}
    The application of machine learning to the study of coronal mass ejections (CMEs) and their impacts on Earth has seen significant growth recently. Understanding and forecasting CME geoeffectiveness is crucial for protecting infrastructure in space and ensuring the resilience of technological systems on Earth. Here we present GeoCME, a deep-learning framework designed to predict, deterministically or probabilistically, whether a CME event that arrives at Earth will cause a geomagnetic storm. A geomagnetic storm is defined as a disturbance of the Earth's magnetosphere during which the minimum Dst index value is less than $-50$ nT. GeoCME is trained on observations from the instruments including LASCO C2, EIT and MDI on board the Solar and Heliospheric Observatory (SOHO), focusing on a dataset that includes 136 halo/partial halo CMEs in Solar Cycle 23. Using ensemble and transfer learning techniques, GeoCME is capable of extracting features hidden in the SOHO observations and making predictions based on the learned features. 
    Our experimental results demonstrate the good performance of GeoCME, achieving a 
    Matthew's correlation coefficient of 0.807 
    and a true skill statistics score of 0.714 when the tool is used as a deterministic prediction model. When the tool is used as a probabilistic forecasting model, it achieves a Brier score of 0.094 and a Brier skill score of 0.493. These results are promising, showing that the proposed GeoCME can help enhance our understanding of CME-triggered solar-terrestrial interactions.
\end{abstract}
\end{frontmatter}

\section{Introduction}

The impacts of geomagnetic storms on Earth have been investigated by many researchers \citep[e.g.,][]{2006JGRA..111.2202W,2007JGRA..112.1206N,
2013SpWea..11..585B,2015AdSpR..55.2745S, 
2018SoPh..293...84A,
2018SoPh..293..107J,
2019SoPh..294..154H,
2019SoPh..294..110W,
2020SoPh..295....7A,
2020SoPh..295...74C,
2021MNRAS.506.1186M,
2022SoPh..297...65B,
2022A&A...665A.110P,
2023SoPh..298...64R,
2023SoPh..298..138Z,
2024MNRAS.527.7298H,
2024SoPh..299...40M}. These storms can affect the accuracy of technological systems, such as satellites and communication systems, that rely on precise measurements of the Earth's magnetic field. 
They can also affect power grids by inducing electrical currents that can damage or disrupt the operation of the grids. In general, geomagnetic storms occur due to the interaction between radiation and plasma released by the Sun into the heliosphere and magnetic fields in the plasma environment near Earth \citep{2006JGRA..111.2202W}. The degree of severity exhibited by a storm is assessed through geomagnetic indices, such as the Kp index (Planetary K-index), the AE (Auroral Electrojet) index, and the Dst (Disturbance Storm Time) index. 
\citet{doi:https://doi.org/10.1002/9781118663837.ch2} discussed the meaning of these indices. Other geomagnetic indices, such as the SYM-H index and the ASY-H index, are similar to the Dst index, but are available in high resolution, with intervals as short as 1 minute or 5 minutes \citep{2006JGRA..111.2202W}. 
  
Coronal mass ejections (CMEs),
which carry strong southward-directed magnetic fields,
may cause intense geomagnetic storms
\citep{2019RSPTA.37780096V,2022A&A...667A.133B, 2023A&A...679A..97M}.
Predicting whether a CME will hit Earth and
when it will reach Earth is a challenging task.
Efforts to tackle this task include the use of
empirical models
\citep[e.g.,][]{1998GeoRL..25.3019B,
2004JGRA..109.6109M,
2005AdSpR..36.2289G},
drag-based models
\citep[e.g.,][]{2002JGRA..107.1019V,2021FrASS...8...58D},
physics-based models
\citep[e.g.,][]{2001JGR...10620985F,2002GeoRL..29.1390M} and
machine learning models
\citep[e.g.,][]{2018ApJ...855..109L,2022FrASS...913345A,
2023ApJ...954..151G,
2023ApJS..268...69Y,
2024ApJ...963..121C},
among others
\citep[e.g.,][]{2014SpWea..12..448Z,2023ApJ...948...78S}.
Machine learning models have also been used to predict the
geoeffectiveness of CMEs.
For example, \citet{2019JASTP.19305036B} adopted logistic regression with numerical CME parameters to
make predictions.  
\citet{2021RemS...13.1738F} presented a deep neural network to predict
the geoeffectiveness and arrival time of CMEs.
The authors used data from SOHO's Large Angle and Spectrometric Coronagraph (LASCO) C2 Field-of-View (FOV) and Extreme Ultraviolet Imaging Telescope (EIT), along with SDO's Atmospheric Imaging Assembly (AIA) observations. 
\citet{2022ApJ...934..176P} explored 
several machine learning methods
such as logistic regression, k-nearest neighbors, and support vector machines, together with solar onset parameters,
to predict the geoeffectiveness of CMEs.

The aforementioned studies
predict CMEs that
reach Earth and cause geomagnetic storms as ``geoeffective,''
and predict all others,
including CMEs that do not reach Earth, 
as ``non-geoeffective.''
In contrast, we focus on
CMEs that arrive at Earth, and predict whether they will cause geomagnetic storms.
Since the problem we attempt to solve here differs
from those addressed in previous work,
the way we collect data for model training and testing
is different from those used in previous work.
The criterion for a disturbance of the Earth's magnetosphere to be
considered a geomagnetic storm
is that its minimum Dst value must be less than $-50$ nT \citep{1994JGR....99.5771G,2022FrASS...9.5880T}.
The Dst index, measured in nanoteslas (nT), is a key indicator used in space weather research to quantify the intensity of geomagnetic storms \citep{doi:https://doi.org/10.1002/9781118663837.ch2}. It reflects the effect of geomagnetic disturbances caused by solar activity on Earth, 
where lower Dst values correspond to stronger storms.

Our work is based on SOHO observations,
including
LASCO C2 and EIT images, as well as Michelson Doppler Imager (MDI) magnetograms.
Our goal is to understand whether machine learning can capture
any possible connection between the SOHO observations and CME geoeffectiveness.
We propose a deep learning framework, named
GeoCME,
to achieve this goal.
Our main assumption is that the CMEs at hand
have already arrived at Earth.
In practice, how do we know whether a CME can reach Earth?
This question can be answered using existing CME arrival prediction methods \citep[e.g.,][]{2016MNRAS.456.1542S,
2018ApJ...855..109L,
2021SpWea..1902553A,
2021FrASS...8...58D,
2021CosRe..59..268K,
2022A&A...667A.133B,
2023ApJ...954..151G,2024ApJ...963..121C}.
Thus, the use of GeoCME is a two-step process.
In the first step,
we use the existing methods mentioned above
to predict whether a CME
would arrive at Earth.
If the CME is predicted to reach Earth, then in the second step
we use GeoCME to predict whether the CME will cause a geomagnetic storm,
i.e., whether the CME is geoeffective.

The remainder of this paper is organized as follows.
Section \ref{sec:data} describes the data used in our study. 
Section \ref{sec:methods} presents the architecture and configuration details of GeoCME. 
Section \ref{sec:results} reports the experimental results. 
Section \ref{sec:conc} presents a discussion and concludes the article.   

\section{Data}
\label{sec:data}

 \begin{figure*}
   \centering
   \includegraphics[width=1\hsize]{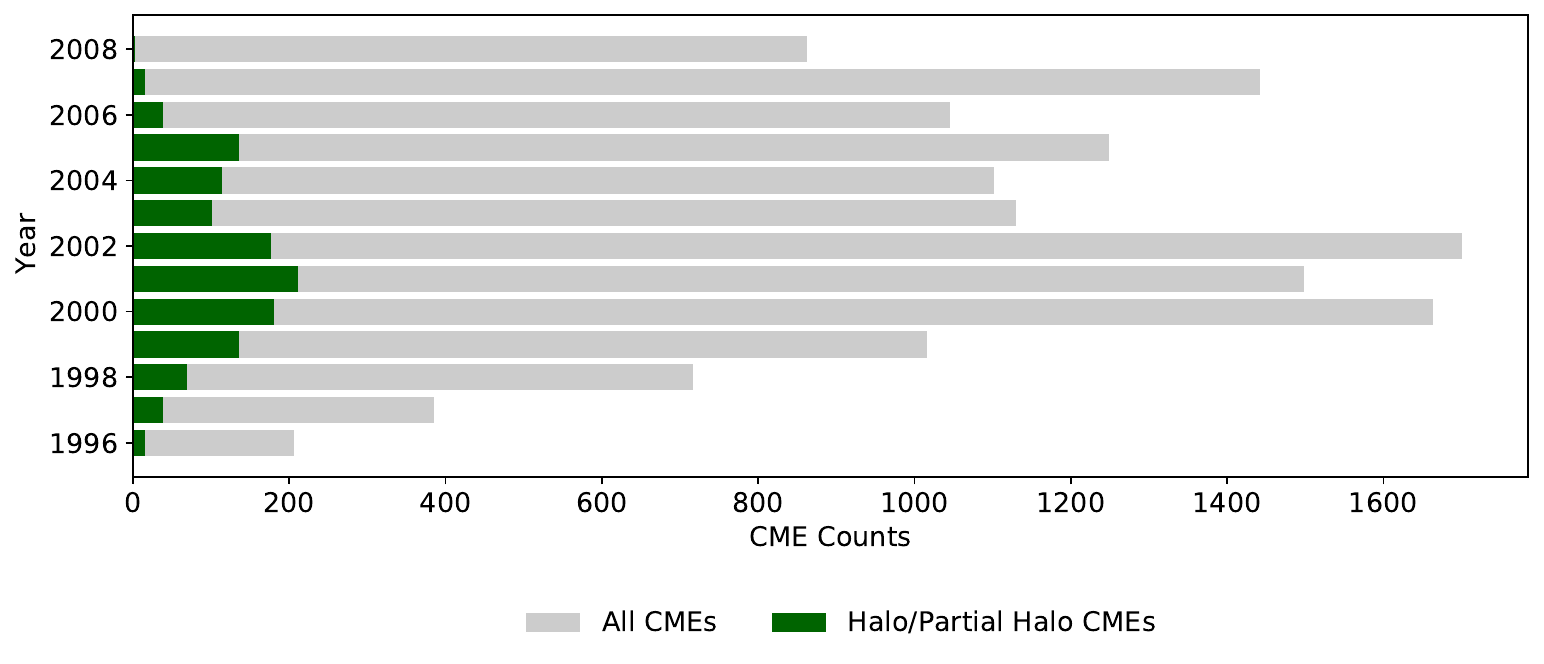}
      \caption{Chart showing the total counts of 
      halo/partial halo CMEs among all CMEs
       during Solar Cycle 23 (1996-2008) according to the SOHO/LASCO CME catalog.}
         \label{Fig:SC23_insights}
   \end{figure*}  

   \begin{figure*}
   \centering
   \includegraphics[width=1\hsize]{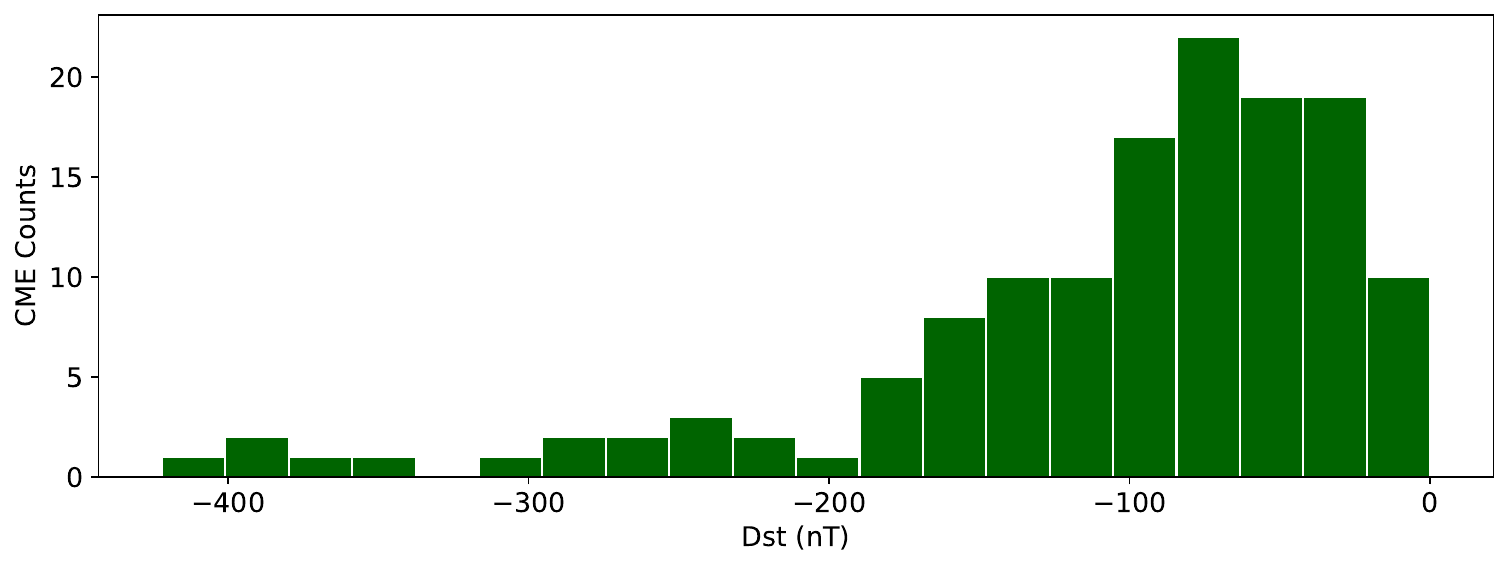}
      \caption{Distribution of the Dst index values caused by the 136 halo/partial halo CME events in our dataset. 
              }
         \label{Fig:dst_distribution}
   \end{figure*}
   
\begin{figure*}
   \centering
   \includegraphics[width=1\hsize]{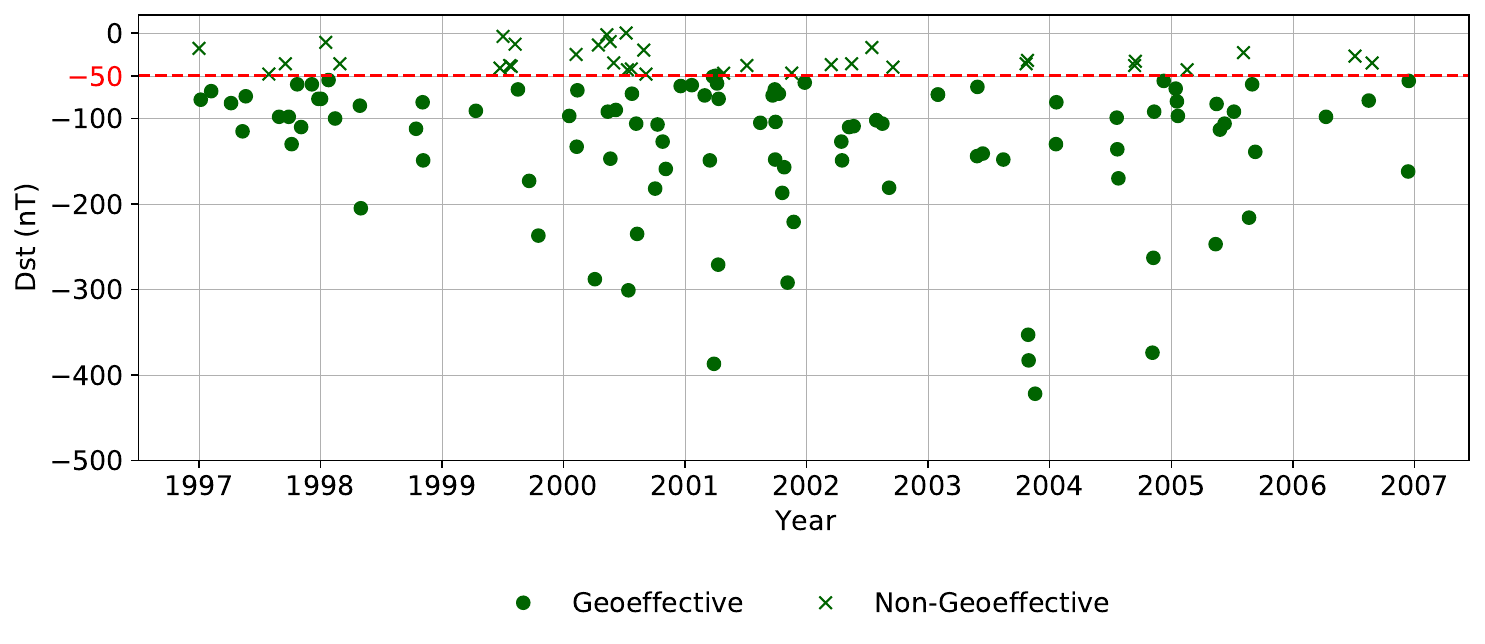}
      \caption{Breakdown analysis of the geoeffective and non-geoeffective CME events in our dataset 
      where a solid circle represents a geoeffective CME event
      and a cross mark represents a non-geoeffective CME event.
 These events were distributed over 10 years, from 1997 to 2006.
              }
         \label{Fig:dstovertime}
   \end{figure*}

We focused on halo/partial halo CMEs in Solar Cycle 23
\citep{2006SpWea...410003M,2007JGRA..112.6112G,2009EP&S...61..595G}.
Figure \ref{Fig:SC23_insights} shows the total counts
of halo/partial halo CMEs in Cycle 23
according to the SOHO/LASCO CME catalog
\citep{2004JGRA..109.7105Y}.
The CME events used in our study were obtained from the list of interplanetary coronal mass ejections (ICMEs), known as the RC list, compiled and maintained by
\citet{2010SoPh..264..189R}. 
We chose 145 CME events within the RC list that occurred in Solar Cycle 23 and arrived at Earth (i.e. with arrival-time data). 
We used the SOHO/LASCO CME catalog
to identify and select
141 halo/partial halo CME events among the 145 CME events.
The RC list shows the minimum Dst index value caused by a CME during its interplanetary interaction with the Earth's magnetosphere. 
We excluded those CME events without Dst index values, which resulted in a total of 136 halo/partial-halo CME events.
 Figure \ref{Fig:dst_distribution} shows the distribution of the minimum Dst values caused by the 136 events.
 As mentioned in the previous section,
 a value of $-50$ nT was used for the Dst index to determine the geoeffectiveness of CMEs \citep{1994JGR....99.5771G,2022FrASS...9.5880T}.
 As a consequence,
 among the 136 halo/partial halo CME events analyzed, 101 were identified as geoeffective, while 35 were classified 
 as non-geoeffective. 
 Figure \ref{Fig:dstovertime} 
 provides a breakdown analysis of the geoeffective and non-geoeffective CME events in our dataset.
 These events were distributed over 10 years, from 1997 to 2006.

\begin{figure*}
       \centering
       \includegraphics[width=1\hsize]{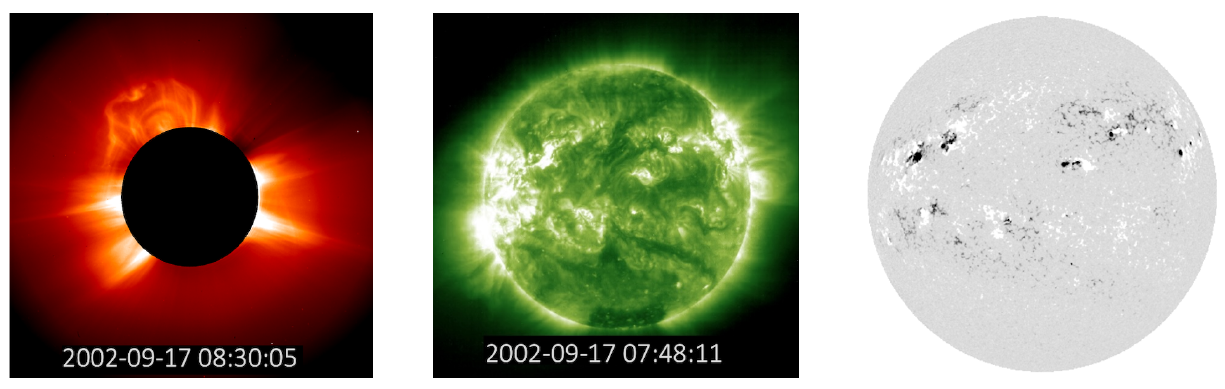}
       \caption{SOHO observations on the CME event 
       that occurred at 08:06:00 UT on 17 September 2002. 
       Shown from left to right are a LASCO C2 image,
       an EIT 195 \AA \hspace*{+0.03cm} image, and a full-disk MDI magnetogram.}
       \label{Fig:C2EITMDI}
    \end{figure*}

When training and testing our GeoCME framework, we used three types of SOHO data, 
namely LASCO C2, EIT 195 \AA, and MDI magnetogram images. 
Figure \ref{Fig:C2EITMDI} shows
the SOHO observations
on the CME event that occurred at 08:06:00 UT on 17 September 2002,
which are, from left to right, LASCO C2, EIT, and MDI, respectively.
LASCO C2 coronagraph captures images of the Sun from 1.5 to 6 R$\odot$ \citep{1995SoPh..162..357B}. 
We constructed base-difference images for LASCO C2 by subtracting the pre-event image (base) from subsequent images of
the event to enhance the visibility of dynamic solar features while minimizing static background information. 
This technique is widely used to improve machine learning of CME image features in LASCO C2 observations \citep{2019ApJ...881...15W, 2023ApJ...958L..34A}.  

Our data collection process follows a systematic approach centered on CME appearance times, as listed in the
SOHO/LASCO CME catalog.
For LASCO C2, we collected images 10 minutes
before a CME event and up to 4 hours after the event \citep{2021RemS...13.1738F}, averaging 12.34 images per event, totaling 1679 images. 
EIT images were collected from 4 hours before the event
to the event time, with an average of 7.59 images per event, resulting in 1033 images.
MDI magnetograms included the last three observations before the event, averaging 2.9 images per event for a total of 395 images. 
In total, this dataset contains 3107 images.

\section{Methodology} 
\label{sec:methods}

\subsection{Transfer Learning}

We addressed the challenge of working with a relatively small dataset with 3107 images by using transfer learning.
The transfer learning approach
involved evaluating the efficacy of several pre-trained deep learning models, including ResNet \citep{DBLP:conf/cvpr/HeZRS16}, InceptionNet \citep{DBLP:conf/cvpr/SzegedyVISW16, DBLP:conf/aaai/SzegedyIVA17}, VGG \citep{DBLP:journals/corr/SimonyanZ14a}, MobileNet \citep{DBLP:conf/cvpr/SandlerHZZC18}, DenseNet \citep{DBLP:conf/cvpr/HuangLMW17}, Xception \citep{DBLP:conf/cvpr/Chollet17}, and EfficientNet \citep{DBLP:conf/icml/TanL19}. 
These pre-trained models were originally designed to
perform representation learning, feature extraction, and image classification.
Our experiments revealed that ResNet152 and InceptionResNetV2,
both pre-trained on the ImageNet dataset \citep{DBLP:conf/cvpr/DengDSLL009}, 
achieved the best results
in geoeffective CME prediction. 

\begin{figure*}
       \centering
       \includegraphics[width=1\hsize]{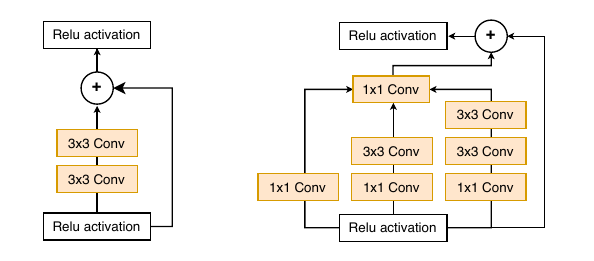}
       \caption{Illustration of a residual block (left)  and an InceptionResNet module (right). The residual block consists of two $3\times3$ convolutional layers followed by a residual connection that allows gradients to flow directly through the network, improving training efficiency. The InceptionResNet module includes parallel convolutional paths with $1\times1$ and $3\times3$ filters, which, combined with a residual connection, capture image features at multiple scales to maintain efficient gradient flow.}
       \label{Fig:Resnet_Inception}
    \end{figure*}

Residual blocks in ResNet152 and inception modules in InceptionResNetV2 are core components that improve the performance of convolutional neural networks. Residual blocks help train very deep networks by allowing gradients to flow more easily through shortcut connections, thus solving the vanishing-gradient problem. Inception modules, on the other hand, use parallel convolutional filters of different sizes to capture image features at multiple scales. 
InceptionResNet combines
residual blocks and inception modules, integrating residual connections
with the inception structure to leverage the strengths of both. Figure \ref{Fig:Resnet_Inception} shows
a residual block \citep{DBLP:conf/cvpr/HeZRS16} in ResNet152 and
an InceptionResNet module \citep{DBLP:conf/cvpr/SzegedyVISW16} in InceptionResNetV2. 
These components improve the accuracy and efficiency of deep neural networks, making them suitable for complex tasks such as recognizing patterns of solar imagery.

Our transfer learning approach, where a pre-trained image classification model is adapted to a new task (i.e., geoeffective CME prediction), provides an effective way to build a new model to specific needs without the substantial training data usually required for complex deep learning models.

\subsection{The Ensemble Model}

\begin{figure*}
   \centering
   \includegraphics[width=1\hsize]{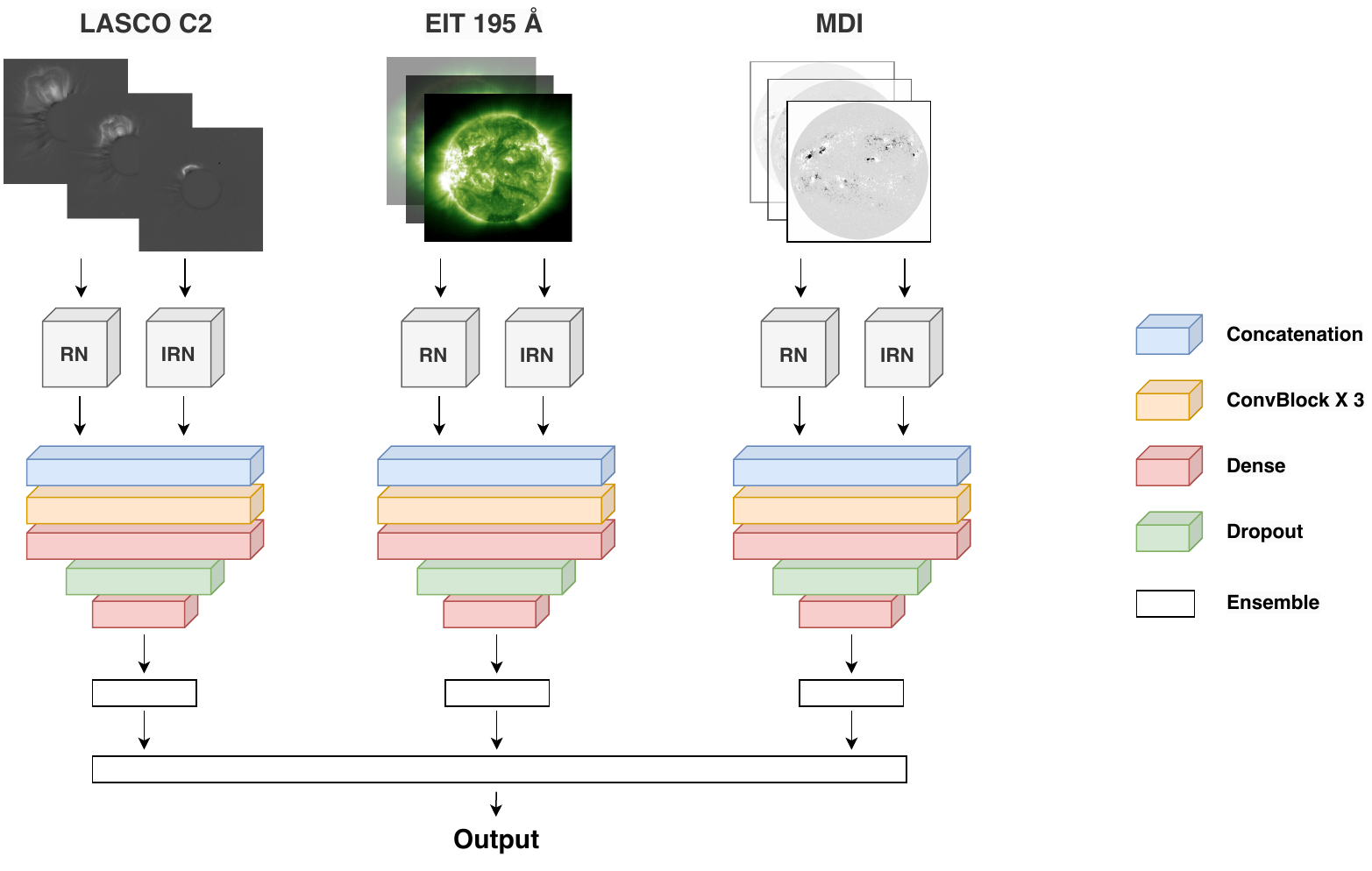}
      \caption{Illustration of the GeoCME architecture. The ensemble model consists of three equal pipelines (left, middle, right), each dedicated to one of the three SOHO instruments (LASCO C2, EIT, and MDI), respectively. Each pipeline begins with two base models, namely ResNet152 (RN) and InceptionResNetV2 (IRN), followed by a concatenation layer that combines the output values of the two base models. This concatenation layer is succeeded by three convolutional blocks, followed by two dense layers with 1024 neurons and 1 neuron, respectively, with a dropout layer between them. Each pipeline ends with an ensemble layer that produces the output of the corresponding SOHO instrument. 
      Finally, the output values of the three pipelines corresponding to the three SOHO instruments are fed to another ensemble layer to produce the final result.}
         \label{Fig:GeoCME}
   \end{figure*}

To further improve
feature extraction capabilities, we combined
ResNet152 and InceptionResNetV2,
referred to as base models,
into an integrated framework (GeoCME).
This ensemble approach aimed to capitalize on the strengths of each
base model, thereby enhancing the overall performance of the
feature extraction process and, subsequently, the accuracy of
geoeffective CME prediction. 
Figure \ref{Fig:GeoCME} illustrates the architecture of the GeoCME framework,
and Table \ref{table:GeoCME_Architecture}
presents its configuration details.
Table \ref{table:pretrained_models_parameters}
summarizes the parameters of the base models
used in GeoCME.

For a given CME event, 
we feed the event's image from each instrument (LASCO C2, EIT, MDI)
into the two base models, ResNet152 (RN) and InceptionResNetV2 (IRN), respectively. 
For each instrument, the output values of the two base models
are fed into a concatenation layer. 
The concatenated features pass through three convolutional blocks (ConvBlock), each equipped with a 2D convolution layer with 64, 128, and 256 filters, respectively.
Each convolutional block has a kernel size of $3 \times 3$, paired with LeakyReLU activation and batch normalization for stability. The features are flattened and processed through a dense layer
of 1024 neurons. 
To avoid overfitting, a dropout layer with a rate of 0.3 is placed
after the dense layer of 1024 neurons.
The dropout layer is followed by another dense layer with 1 neuron.

As shown in Figure \ref{Fig:GeoCME}, each SOHO instrument
(LASCO C2, EIT, MDI) uses the pipeline described above to produce
one prediction per image. 
For a CME event with multiple images from the same instrument, an ensemble layer calculates the mean of all output values of the images from the same instrument for the CME event. At this stage, we have one predicted value per instrument for the CME event. The final ensemble layer then calculates the mean of the three predicted values
from the three instruments to get the final output for the
CME event.
This final output is the probability that the CME event will be geoeffective, i.e. the CME event will cause a geomagnetic storm.
We implemented a threshold of
0.6 in the output layer
to obtain a deterministic model.
If the probability is greater than or equal to the threshold,
then the GeoCME model predicts that the CME event is geoeffective;
otherwise, the model predicts that
the CME event is non-geoeffective.

\begin{table}
\caption{Configuration details of the GeoCME framework.}
\label{table:GeoCME_Architecture}
\begin{tabular}{lccccc}
\hline
Layer & Kernel No. & Kernel Size & Regularization & Activation & Output \\
\hline
ConvBlock 1 & 64 & $3\times3$ & Batch Norm & LeakyReLU & 
$8\times8\times64$\\
ConvBlock 2 & 128 & $3\times3$ & Batch Norm & LeakyReLU  & 
$8\times8\times128$\\
ConvBlock 3 & 256 & $3\times3$ & Batch Norm & LeakyReLU  & 
$8\times8\times256$\\
Dense 1 & -- & -- & Batch Norm & LeakyReLU  & 1024 \\
Dense 2 & -- & -- & -- & Sigmoid  & 1 \\
\hline
\end{tabular}
\end{table}

\begin{table}
\caption{Base model parameters.}
\label{table:pretrained_models_parameters}
\begin{tabular}{lcc}
\hline
Base Model & Layer Number & Parameter Number \\
\hline
ResNet152 & 152 & 58.50M \\
InceptionResNetV2 & 164  & 54.39M \\
\hline
\end{tabular}
\end{table}

\begin{table}
\caption{Hyperparameters for GeoCME training.}
\label{table:GeoCME_Hyperparameters}
\begin{tabular}{cccccc}
\hline
Loss Function & Optimizer & Dropout Rate & Batch Size & Epochs \\
\hline
WBCE & Adam & 0.3 & 32 & 100 \\
\hline
\end{tabular}
\end{table}
We have an imbalanced dataset at hand, which contains
a positive (or majority) class
with 101 geoeffective CME events and
a negative (or minority) class with 35 non-geoeffective CME events.
We use
the Weighted Binary Cross Entropy (WBCE) loss
function to combat the imbalance issue
within the dataset
\citep{Goodfellow-et-al-2016, 2020ApJ...890...12L, 2022ApJS..260...16A}.  
Let $N$ denote the number of events in the training or validation set.
Let $w_0$ denote the weight for the negative (or minority) class and
let $w_1$ denote the weight for the positive (or majority) class.
The weight assignment is based on the ratio of sizes between the majority and minority classes, with a higher weight assigned to the minority class, as shown below.
\begin{equation}
\mbox{WBCE} = -\frac{1}{N} \sum_{i=1}^{N} \left[ w_0 y_i \log(\hat{y}_i) + w_1 (1 - y_i) \log(1 - \hat{y}_i) \right].
\end{equation}
Here, $y_i$ denotes the label of the $i$th event, with $y_i = 1$ for a geoeffective CME and $y_i = 0$ for a non-geoeffective CME, and $\hat{y}_i$ represents the predicted probability for the $i$th event being positive. The WBCE method ensures that the minority class
is emphasized more in the loss calculation, effectively addressing the imbalance issue in our dataset.
During model training,
we use adaptive moment estimation (Adam) as optimizer \citep{Goodfellow-et-al-2016},
with a batch size of 32, and a total of 100 epochs.
Table \ref{table:GeoCME_Hyperparameters} summarizes the hyperparameters used in model training. 
These hyperparameter values are obtained by using the grid search capability from the Python machine learning 
library, scikit-learn \citep{10.5555/1953048.2078195}.
The validation set used for tuning the hyperparameters is described below.

\section{Results} \label{sec:results}

\subsection{Experimental Setup} \label{sec:setup}
\begin{figure}
   \centering
   \includegraphics[width=1\hsize]{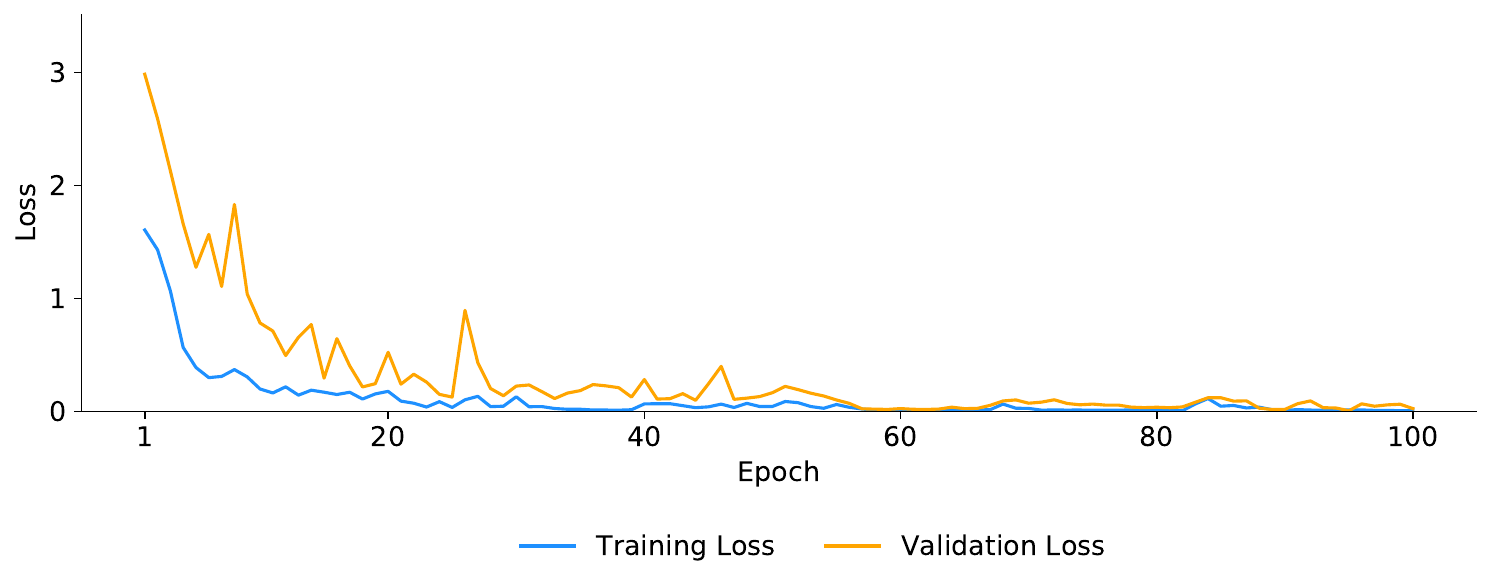}
      \caption{Training and validation learning curves showing GeoCME is a well-fit model for geoeffective CME prediction.} 
         \label{Fig:learning_curve}
   \end{figure}

We adopted an 80:20 scheme to train and test the GeoCME framework.
Specifically, we used
80\% of the CME events from each of the ``geoeffective'' 
and ``non-geoeffective'' classes for model training and used
the remaining 20\% of the events from each class
for model testing.
Furthermore, we allocated 10\% of the training data for each class for validation, so that the performance of our model was regularly evaluated against unseen data throughout the training process. 
Figure \ref{Fig:learning_curve} presents
the GeoCME training and validation
learning curves.
The downward and convergence trends in
the learning curves
demonstrate the effectiveness of GeoCME learning
and its capacity to generalize successfully
to new data.
We note that the two base models of
GeoCME (ResNet152 and InceptionResNetV2)
are pre-trained on the extensive ImageNet dataset. 
We used a relatively small amount of new training data to retrain the two complex
models for our use through transfer learning.
Because the complex models have been well pre-trained and GeoCME is a fusion of them, we see that the learning curves of GeoCME converge well in
Figure \ref{Fig:learning_curve}.

In the experimental study,
we adopted two metrics to evaluate GeoCME's performance: Matthew's Correlation Coefficient (MCC) and True Skill Statistics (TSS).
Given a CME event $E$,
we define $E$ as
a true positive (TP)
if the model predicts $E$ as positive (i.e. geoeffective)
and $E$ is indeed positive.
We define $E$ as a true negative (TN)
if the model predicts $E$ as negative
(i.e. non-geoeffective)
and $E$ is indeed negative.
We say that $E$ is a false positive (FP)
if the model predicts $E$ as positive
while $E$ is actually negative;
$E$ is a false negative (FN)
if the model predicts $E$ as negative
while $E$ is actually positive.
When the context is clear, we also use
TP (TN, FP, and FN, respectively) to represent
the total number of true positives
(true negatives, false positives, and false negatives,
respectively) produced by the model.
The MCC and TSS are defined as follows
\citep{2019ApJ...877..121L, 2022ApJS..260...16A}:

\begin{equation}
\mbox{MCC} = \frac{\mbox{TP} \times \mbox{TN} - \mbox{FP} \times \mbox{FN}}{\sqrt{(\mbox{TP} + \mbox{FP})(\mbox{TP} + \mbox{FN})(\mbox{TN} + \mbox{FP})(\mbox{TN} + \mbox{FN})}},
\end{equation}

\begin{equation}
\mbox{TSS} = \frac{\mbox{TP}}{\mbox{TP} + \mbox{FN}} - \frac{\mbox{FP}}{\mbox{FP} + \mbox{TN}}.
\end{equation}

\begin{figure*}
   \centering
   \includegraphics[width=1\hsize]{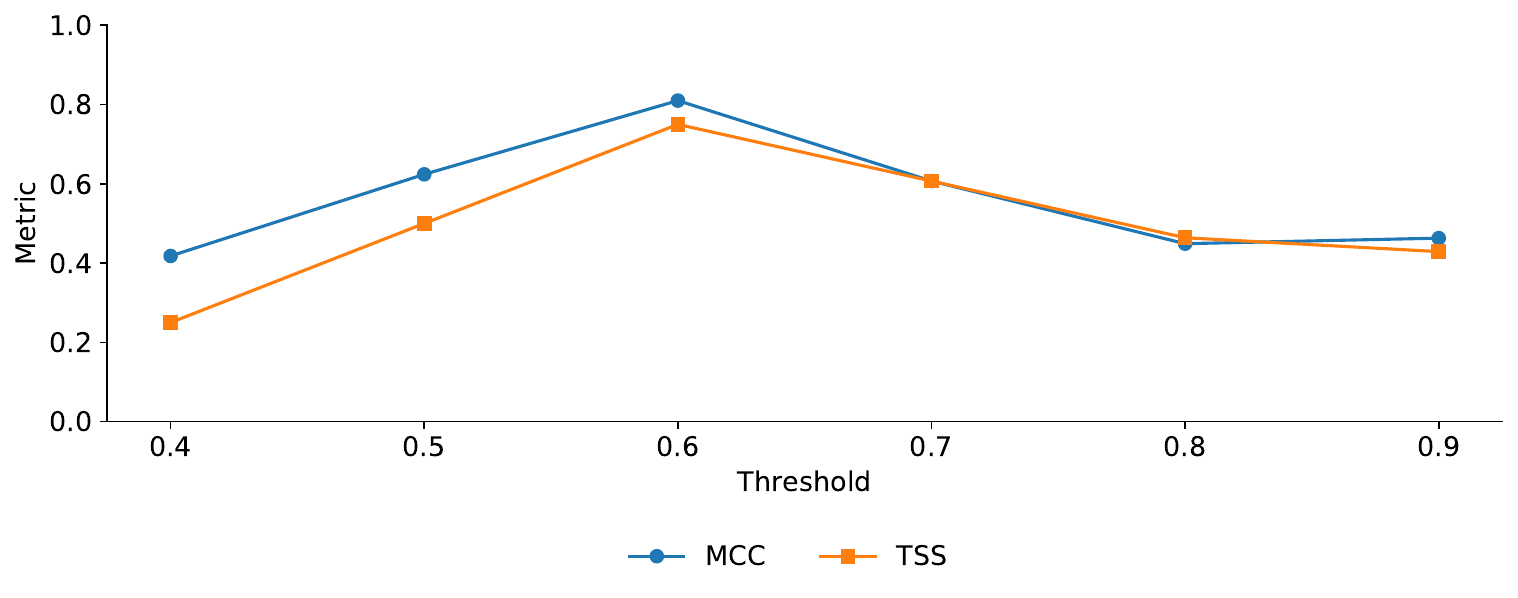}
      \caption{GeoCME's metric values for varying thresholds 
      based on the validation set. 
      The best metric values are obtained when the threshold is set to 0.6.}
         \label{Fig:threshold}
   \end{figure*}

As mentioned above,
we implemented a threshold
in the GeoCME output layer
to obtain a deterministic prediction model.
If the probability produced by GeoCME for a given
CME event is greater than or equal to the threshold,
then the model predicts that the CME event is geoeffective;
otherwise, the model predicts that
the CME event is non-geoeffective.
Figure \ref{Fig:threshold} presents GeoCME's metric values  
for varying thresholds based on the validation set.
The best metric values are obtained when the threshold is set to 0.6.
As a consequence, we used the 0.6 threshold in our study.

\subsection{Performance Evaluation} \label{sec:evaluation_deterministic}

\begin{figure*}
   \centering
   \includegraphics[width=1\hsize]{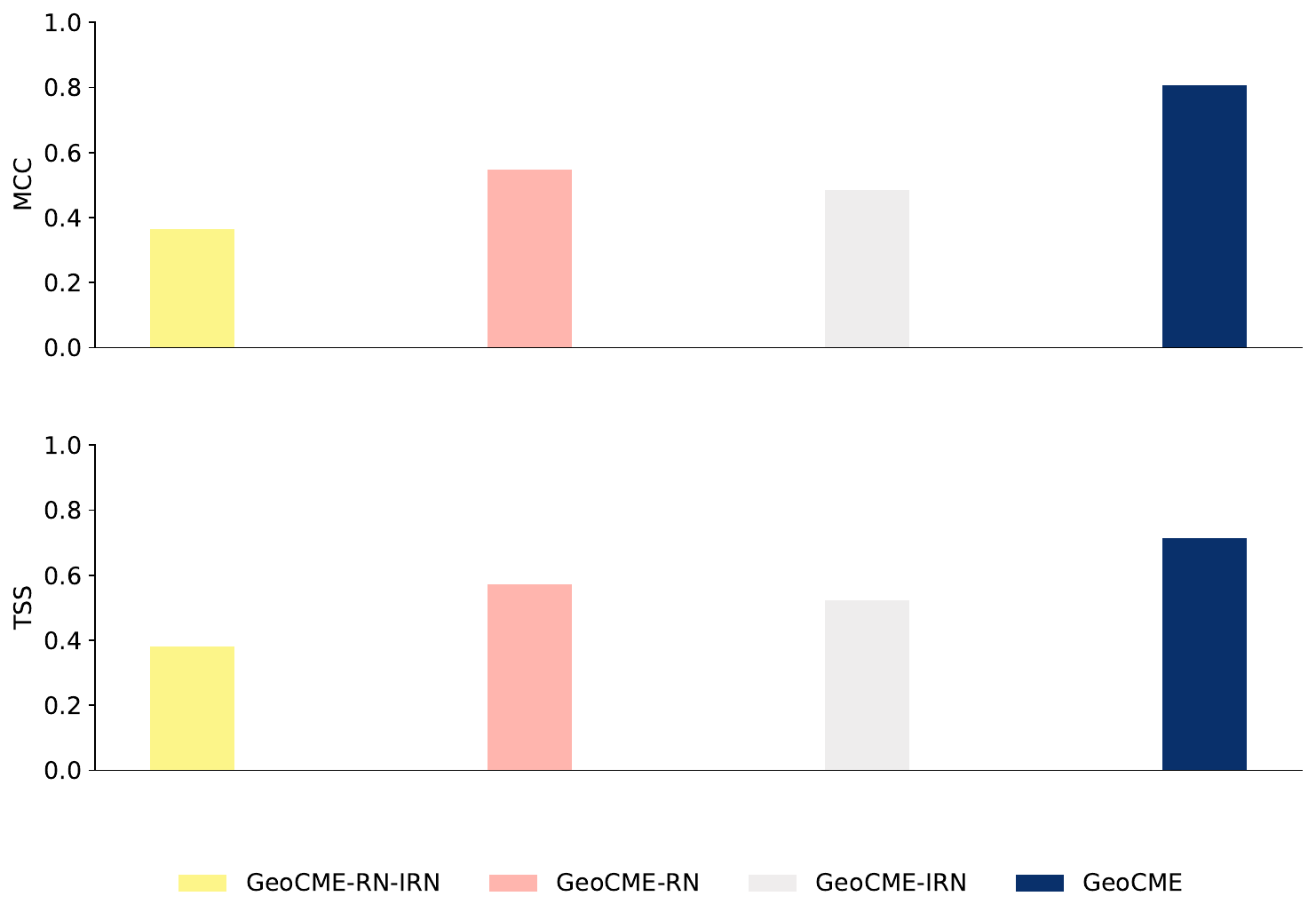}
      \caption{Results of the ablation tests for assessing four networks 
      (GeoCME-RN-IRN, GeoCME-RN, GeoCME-IRN, and GeoCME)
      used as deterministic prediction models
      where GeoCME-RN-IRN represents GeoCME with RN and IRN models removed, GeoCME-RN represents GeoCME with RN models removed, GeoCME-IRN represents GeoCME with IRN models removed, and GeoCME represents the full model. 
      (Top) MCC of the networks tested.
      (Bottom) TSS of the networks tested.
      GeoCME achieves the best performance among all tested networks.}
         \label{Fig:model_ablation_deterministic}
   \end{figure*}

We conducted ablation tests to analyze and
evaluate the components of our GeoCME framework.
GeoCME contains two pre-trained base models
(see Figure \ref{Fig:GeoCME}):
ResNet152 (RN) and InceptionResNetV2 (IRN).
We considered three variants of GeoCME:
GeoCME-RN-IRN,
GeoCME-RN,
GeoCME-IRN.
GeoCME-RN-IRN denotes GeoCME with the RN and IRN models removed.
This subnet contains only the inherent structure of GeoCME
without the pre-trained models used for feature extraction.
Thus, there is no transfer learning in GeoCME-RN-IRN.
GeoCME-RN denotes GeoCME with the RN models removed.
GeoCME-IRN denotes GeoCME with the IRN models removed.
Figure \ref{Fig:model_ablation_deterministic} compares the performance
of the four networks
used as deterministic prediction models. 

It can be seen in
Figure \ref{Fig:model_ablation_deterministic} that
the variants (GeoCME-RN, GeoCME-IRN, and GeoCME-RN-IRN), each missing a key component or more, achieved varied performance levels. 
The GeoCME framework, which integrates all its components, shows the best performance by achieving the highest MCC of 0.807 and the highest TSS of 0.714. 
GeoCME-RN-IRN, which lacks both the ResNet and InceptionResNet base models, exhibits the most significant drop in prediction accuracy, with the lowest
MCC of 0.365 and the lowest TSS of 0.380. This highlights the impact of excluding transfer learning on
GeoCME's performance.
Furthermore, GeoCME-RN performs better than GeoCME-IRN, emphasizing the importance of InceptionResNet in
improving the prediction accuracy.

\begin{figure}
\centering
   \includegraphics[width=0.5\hsize]{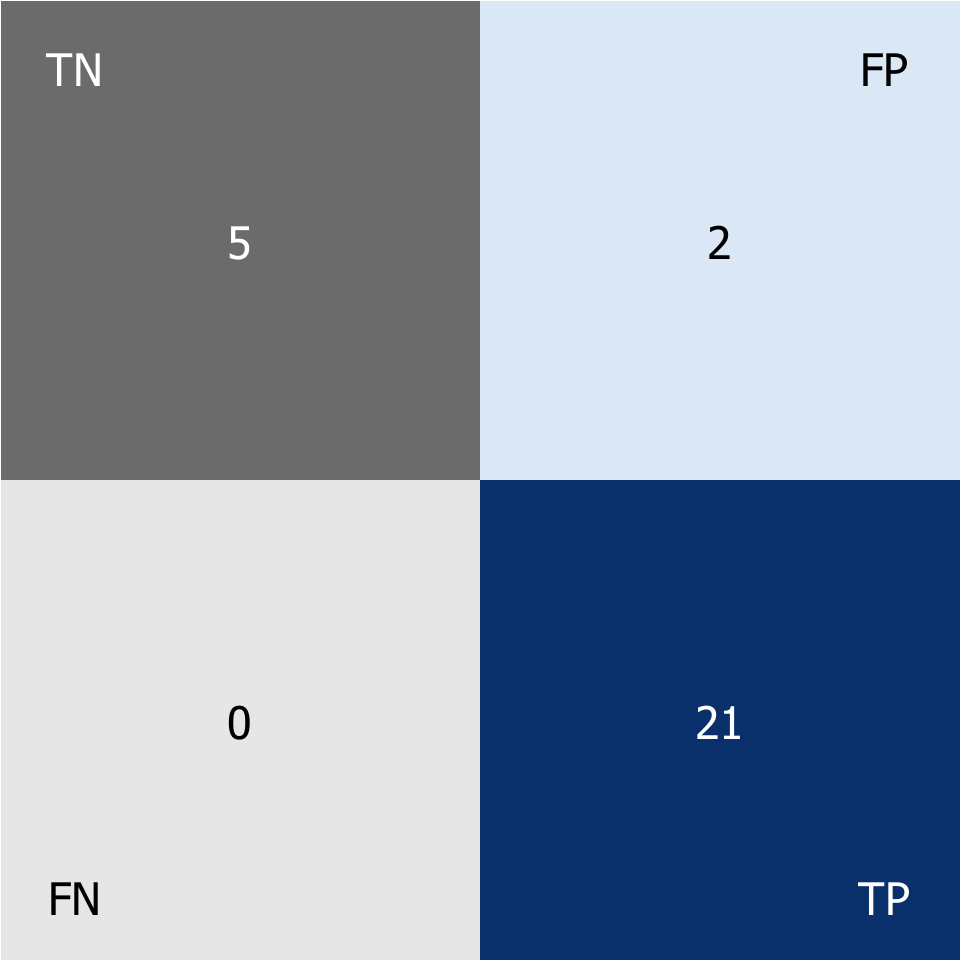}
     \caption{The confusion matrix obtained by GeoCME 
     used as a deterministic prediction model
     on the test set.
             }
        \label{Fig:conf_matrix}
   \end{figure}

Figure \ref{Fig:conf_matrix} presents the confusion matrix
obtained by GeoCME,
which provides a breakdown analysis of errors that occur when
the model makes predictions in the test set.
There are 28 CME events in the test set.
Approximately (21+2)/28 = 82\% of the events in the test set are predicted to be geoeffective.
Approximately 2/(21+2) = 8.7\% of the predictions are false alarms (false positives).
The model's FP value is 2,
indicating that it is a relatively sensitive model in the sense that it predicts two CME events
as positive, while
these events do not cause geomagnetic storms.
However, the model does not miss any geomagnetic storms, as reflected by the fact that the model's FN value is zero.

\begin{figure*}
   \centering
   \includegraphics[width=1\hsize]{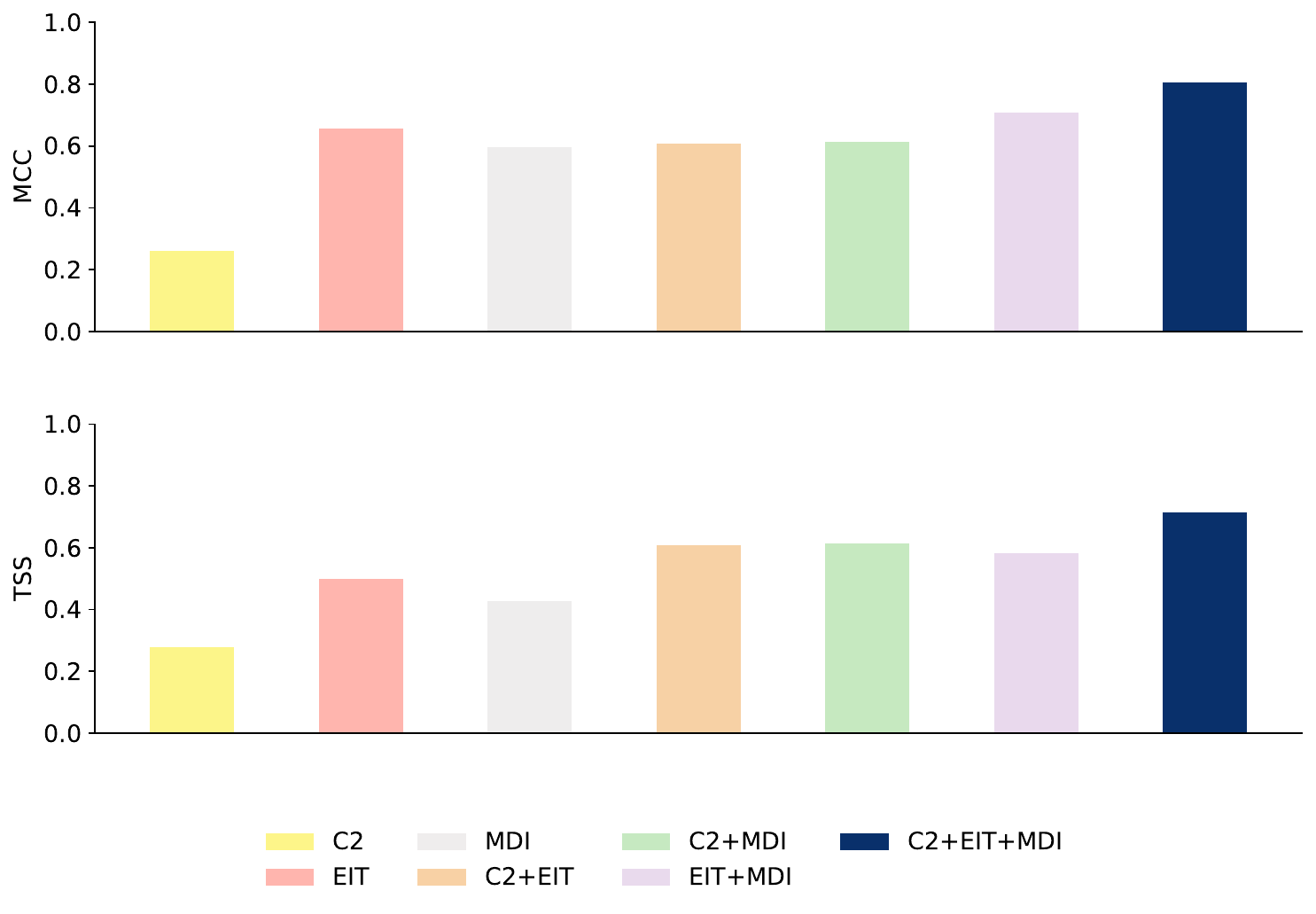}
      \caption{Results of the ablation tests for assessing seven cases 
      (C2, EIT, MDI, C2+EIT, C2+MDI, EIT+MDI, C2+EIT+MDI)
      using GeoCME as a deterministic prediction model
      where C2 represents the LASCO C2 images,
      EIT represents the EIT images,
      MDI represents the MDI magnetogram images,
      C2+EIT represents the combination of LASCO C2 and EIT images,
      C2+MDI represents the combination of LASCO C2 and MDI images,
      EIT+MDI represents the combination of EIT and MDI images,
      and C2+EIT+MDI represents the combination of LASCO C2, EIT and MDI images.
      (Top) MCC of the seven cases tested.
      (Bottom) TSS of the seven cases tested.
      C2+EIT+MDI achieves the best performance among all tested cases.}
         \label{Fig:data_ablation_deterministic}
   \end{figure*}

Each CME event $E$ is accompanied by images from three distinct SOHO instruments
(LASCO C2, EIT and MDI).
To evaluate the effectiveness of these images,
we conducted additional experiments in which we considered the following seven cases.
\begin{itemize}
\item 
$E$ has only LASCO C2 images (denoted C2).
\item 
$E$ has only EIT images (denoted EIT).
\item 
$E$ has only MDI magnetogram images (denoted MDI).
\item 
$E$ has only LASCO C2 and EIT images (denoted C2+EIT).
\item 
$E$ has only LASCO C2 and MDI images (denoted C2+MDI).
\item 
$E$ has only EIT and MDI images (denoted EIT+MDI).
\item 
$E$ has all the images of the three instruments (denoted C2+EIT+MDI).
\end{itemize}
For each case, we custom-built GeoCME to use the provided data. 
Figure \ref{Fig:data_ablation_deterministic}
presents the MCC and TSS results for the seven cases
using GeoCME as a deterministic prediction model.
Note that the C2+EIT+MDI case in Figure \ref{Fig:data_ablation_deterministic} 
is equivalent to GeoCME in
Figure \ref{Fig:model_ablation_deterministic}.
It can be seen in Figure \ref{Fig:data_ablation_deterministic} 
that the combination of LASCO C2, EIT, and MDI images produces
the most accurate results with a MCC of 0.807 and a TSS of 0.714
as also shown in Figure \ref{Fig:model_ablation_deterministic}, 
indicating that the use of the three types of images together leads to the best performance. 
When the three types of data are used individually and separately, EIT produces the best results, 
with a MCC of 0.657 and a TSS of 0.50, followed by MDI, and LASCO C2 is the least effective.

\subsection{Probabilistic Forecasting}

Our proposed GeoCME can be
easily converted from a deterministic prediction model to a
probabilistic forecasting model as follows. 
Instead of comparing
the probability (ranging from 0 to 1) produced by the GeoCME model with a pre-determined threshold
(which is set to 0.6 in our work), the model simply
outputs the probability. 
For a given CME event,
this output now represents a probabilistic estimate of how
likely the event will be geoeffective, that is, how likely it will cause a geomagnetic storm
with the minimum Dst value less than $-50$ nT.

We use the Brier score \citep[BS;][]{brier1950verification}
 and the Brier skill score \citep[BSS;][]{wilks2010sampling} to assess the performance of a model. 
The Brier score quantifies the accuracy of the probabilistic forecasts
produced by the model
by calculating the squared difference between the predicted probabilities and the actual outcomes.
Mathematically, the Brier score is calculated by the following formula:
\begin{equation}
\mbox{BS} = \frac{1}{N}\sum_{i=1}^{N}(y_i - \hat{y}_i)^2,
\end{equation}
where \(N\) is the number of 
CME events in the test set,
\(y_i\) is the actual outcome for the $i$th event
(with 1 representing ``geoeffective'' and 
0 representing ``non-geoeffective''), 
and \(\hat{y}_i\) is the predicted probability
for the $i$th event. 
BS values range from 0 to 1, with a
perfect score of 0.

The Brier skill score provides a measure of the model's skill relative to a baseline prediction, calculated as:
\begin{equation}
\mbox{BSS} = 1 - \frac{\mbox{BS}}{\frac{1}{N}\sum_{i=1}^{N}(y_i - \bar{y})^2},
\end{equation}
where \(\bar{y}\) = $\frac{1}{N}\sum_{i=1}^{N} y_i$ represents the average of the
actual outcomes
for the events in the test set. 
BSS values range from minus
infinity to 1, with the perfect score being 1.
A BSS of 0 indicates that the model has the same accuracy as the baseline model
and a negative BSS indicates that
the model performs worse
than the baseline.

\begin{figure*}
   \centering
   \includegraphics[width=1\hsize]{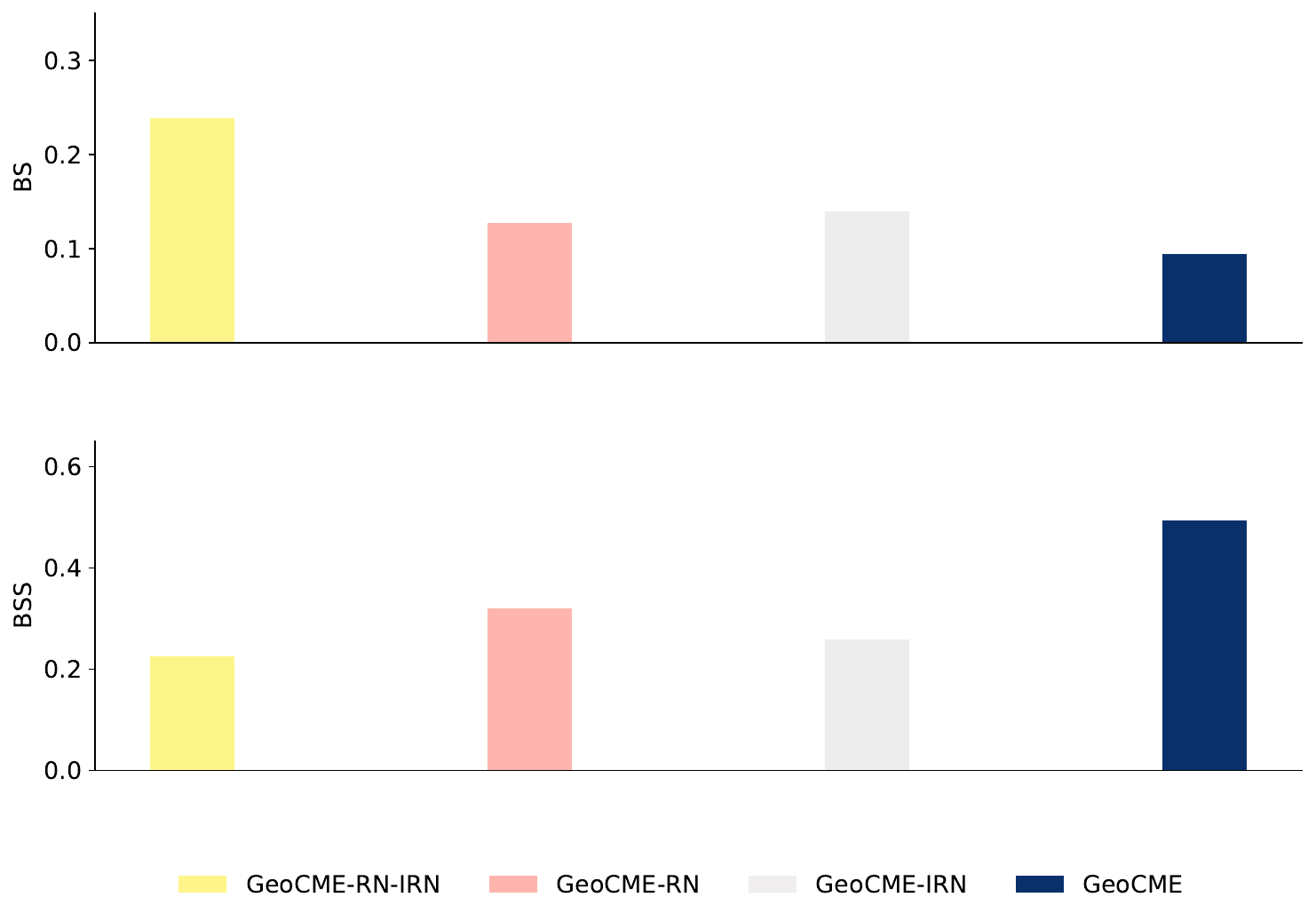}
      \caption{Results of the ablation tests for assessing four networks 
      (GeoCME-RN-IRN, GeoCME-RN, GeoCME-IRN, and GeoCME)
      used as probabilistic forecasting models
      where GeoCME-RN-IRN represents GeoCME with RN and IRN models removed, GeoCME-RN represents GeoCME with RN models removed, GeoCME-IRN represents GeoCME with IRN models removed, and GeoCME represents the full model. 
      (Top) BS of the networks tested.
      (Bottom) BSS of the networks tested.
      GeoCME achieves the best performance among all tested networks.}
         \label{Fig:model_ablation_probabilistic}
   \end{figure*}

Figure \ref{Fig:model_ablation_probabilistic}
compares the four networks,
namely GeoCME-RN-IRN, GeoCME-RN, GeoCME-IRN, and GeoCME,
described in Section \ref{sec:evaluation_deterministic}
where the four networks are now used as probabilistic
forecasting models.
It can be seen in Figure \ref{Fig:model_ablation_probabilistic} that
the GeoCME model again performs the best, achieving the lowest BS of 0.094 and the highest BSS of 0.493. 
GeoCME-RN-IRN, in which both ResNet and InceptionResNet were removed, performs the worst, as reflected by the highest
BS of 0.239 and the lowest BSS of 0.225. 
These results are consistent with those shown in
Figure \ref{Fig:model_ablation_deterministic}
where the four networks were used as deterministic models.

\begin{figure*}
   \centering
   \includegraphics[width=1\hsize]{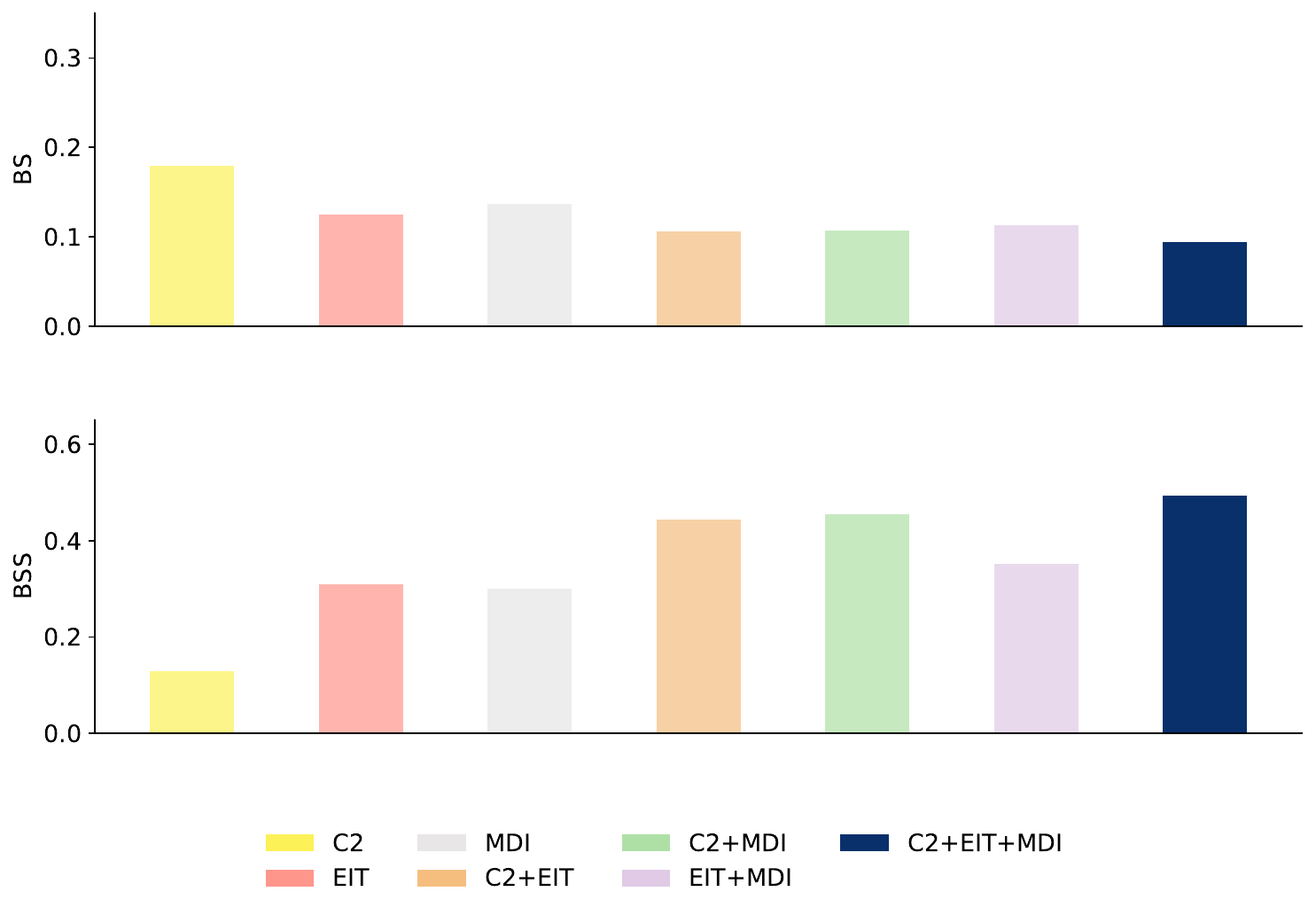}
      \caption{Results of the ablation tests for assessing seven cases 
      (C2, EIT, MDI, C2+EIT, C2+MDI, EIT+MDI, C2+EIT+MDI)
      using GeoCME as a probabilistic forecasting model
      where C2 represents the LASCO C2 images,
      EIT represents the EIT images,
      MDI represents the MDI magnetogram images,
      C2+EIT represents the combination of LASCO C2 and EIT images,
      C2+MDI represents the combination of LASCO C2 and MDI images,
      EIT+MDI represents the combination of EIT and MDI images,
      and C2+EIT+MDI represents the combination of LASCO C2, EIT and MDI images.
      (Top) BS of the seven cases tested.
      (Bottom) BSS of the seven cases tested.
      C2+EIT+MDI achieves the best performance among all tested cases.
              }
         \label{Fig:data_ablation_probabilistic}
   \end{figure*}

Figure \ref{Fig:data_ablation_probabilistic} presents the BS and BSS results for the
seven cases
(C2, EIT, MDI, C2+EIT, C2+MDI, EIT+MDI, C2+EIT+MDI)
defined in Section \ref{sec:evaluation_deterministic},
this time using GeoCME as a probabilistic forecasting model.
It can be seen in Figure \ref{Fig:data_ablation_probabilistic} 
that the combination of LASCO C2, EIT, and MDI images again produces the most accurate results with a BS of 0.094 and a BSS of 0.493, indicating that
the use of all data from the three instruments together achieves the best performance. 
When the three types of data are used individually and separately, EIT yields the best results, 
with a BS of 0.125 and a BSS of 0.310, followed by MDI, and LASCO C2 is the least effective.
These findings are consistent with those shown in
Figure \ref{Fig:data_ablation_deterministic}
where GeoCME was used as a deterministic prediction model.

\section{Discussion and Conclusion} \label{sec:conc}
We presented GeoCME, a deterministic model that employs ensemble and transfer learning techniques to predict whether a CME event reaching Earth will be geoeffective.
Here, a geomagnetic storm is defined as a disturbance of the Earth's magnetosphere during which the minimum value of the Dst index is less than $-50$ nT.
Moreover, we converted the deterministic model to a probabilistic forecasting model,
which estimates the probability that a CME event will be geoeffective.
The GeoCME framework used LASCO C2, EIT 195 \AA, and MDI magnetogram images
collected by SOHO to make predictions.
Our experiments showed that the GeoCME framework can capture the hidden relationships
between the SOHO observations and the CME geoeffectiveness,
achieving reasonably good performance.
Specifically, when used as a deterministic prediction model, GeoCME
achieves a MCC of 0.807 and a TSS of 0.714.
Approximately 82\% of the events in the test set are predicted to be geoeffective.
Approximately 8.7\% of the predictions are false alarms.
When used as a probabilistic forecasting model,
GeoCME achieves a BS of 0.094 and a BSS of 0.493.
Our experiments also showed that using all three types of solar image together (LASCO C2, EIT, and MDI) performs better than using one or two types of solar image.

We adopted an 80:20 scheme in our dataset that covers CME events from
1997 to 2006 for model training and testing, as described in
Section \ref{sec:setup}.
In additional experiments, we conducted a five-fold cross-validation to further
evaluate the GeoCME framework. 
Specifically, we divide the dataset into five equally sized subsets or folds,  
where
every two folds have roughly the same number of geoeffective (non-geoeffective, respectively) CMEs. 
There are 101 geoeffective CMEs and 35 non-geoeffective CMEs in the dataset. 
Thus, each fold contains approximately
20 geoeffective CMEs and 7 non-geoeffective CMEs.
In each run, one fold is used as the test set and the union of the other four folds is used as the training set. 
There are five folds, and hence five runs.
We calculate the average metric values for the five runs.
The five-fold cross-validation process yields  
an average 
MCC of 0.782 
and an average TSS of 0.673
when GeoCME is used as a deterministic prediction model, 
and an average BS of 0.107 and an average BSS of 0.461 when GeoCME is used as a probabilistic prediction model.
Furthermore, in terms of the average metric values,
GeoCME outperforms its subnets (GeoCME-RN-IRN, GeoCME-RN, and GeoCME-IRN)
and performs the best when all three types of solar image together
(LASCO C2, EIT, and MDI)
are used in model training and testing.
These results are consistent with those obtained from the 80:20 scheme.

Our work relies on existing methods
\citep[e.g.,][]{2016MNRAS.456.1542S,
2018ApJ...855..109L,
2021SpWea..1902553A,
2021FrASS...8...58D,
2021CosRe..59..268K,
2022A&A...667A.133B,
2023ApJ...954..151G,2024ApJ...963..121C}
to predict whether a CME event would arrive at Earth.
When a CME event is predicted to arrive at Earth, we then use the proposed GeoCME to predict whether the CME event will be geoeffective, that is, whether it will cause a geomagnetic storm. 
Unlike other studies
\citep{2019JASTP.19305036B,2022ApJ...934..176P},
which used CME or solar onset parameters, GeoCME uses
solar images to make predictions.
The input of GeoCME is composed of directly observed images, which avoids the sophisticated calculation of parameters.
Thus, GeoCME has the potential for operational utilization.
On the basis of our experimental results,
we conclude that GeoCME is a feasible tool
for predicting geoeffective CMEs,
deterministically or probabilistically.
\   \\

\noindent
{\scriptsize
{\bf Acknowledgments}
 The authors thank the members of the Institute for Space Weather Sciences for fruitful discussions.
The CME catalog used in this work was created and maintained at the CDAW Data Center by NASA and the Catholic University of America in cooperation with the Naval Research Laboratory. 
SOHO is an international cooperation project between ESA and NASA. 
}
 
\begin{authorcontribution}
J.W. and H.W. conceived the study.
K.A. wrote the manuscript.
All the authors reviewed the manuscript.
\end{authorcontribution}

\begin{fundinginformation}
K.A. is supported by King Saud University, Saudi Arabia.
J.W. and H.W. acknowledge support from NSF grants AGS-2149748, AGS-2228996, 
OAC-2320147, 
and NASA grants
80NSSC24K0843 and 80NSSC24M0174.
J.J. acknowledges support from NSF grants
AGS-2149748 and AGS-2300341.
V.Y. acknowledges support from NSF grants
AST-2108235,
AGS-2114201,
AGS-2300341, and
AGS-2309939.
\end{fundinginformation}

\begin{dataavailability}
CME events used in this study were compiled from the RC list in
\url{https://izw1.caltech.edu/ACE/ASC/DATA/level3/icmetable2.htm}
and the SOHO/LASCO CME catalog in
\url{https://cdaw.gsfc.nasa.gov/CME_list/}.
SOHO images were collected from the
SOHO science archive at
\url{https://soho.nascom.nasa.gov/data/archive/}.
\end{dataavailability}

\begin{ethics}
\begin{conflict}
The authors declare no conflict of interest.
\end{conflict}
\end{ethics}
  
\bibliographystyle{spr-mp-sola} 

\end{document}